\documentclass[letterpaper, 10 pt, conference]{ieeeconf}  % Comment this line out if you need a4paper

\IEEEoverridecommandlockouts                              % This command is only needed if 
\usepackage{cite}
\usepackage{multicol}
\usepackage{graphicx}
\usepackage{cuted}
\usepackage{capt-of}
\usepackage{amsmath,amssymb}
\usepackage{tcolorbox}
\usepackage{algorithm}
\usepackage{algpseudocode}
\usepackage[bookmarks=true]{hyperref}
\usepackage{subfiles} % Best loaded last in the preamble

\title{\LARGE \bf R2BC: Multi-Agent Imitation Learning from Single-Agent Demonstrations
}

\author{
Connor Mattson$^{1}$, Varun Raveendra$^{1}$, Ellen Novoseller$^{2}$, Nicholas Waytowich$^{2}$, \\ Vernon J. Lawhern$^{2}$, Daniel S. Brown$^{1}$ %
\thanks{$^{1}$Kahlert School of Computing, University of Utah, Salt Lake City, USA }%
\thanks{$^{2}$DEVCOM Army Research Laboratory, Aberdeen Proving Ground, USA}%
}

\begin{document}

\maketitle
\thispagestyle{empty}
\pagestyle{empty}

%%%%%%%%%%%%%%%%%%%%%%%%%%%%%%%%%%%%%%
% TEASER FIGURE
%%%%%%%%%%%%%%%%%%%%%%%%%%%%%%%%%%%%%%
\begin{strip}\centering
\vspace{-5em}
\includegraphics[width=\textwidth]{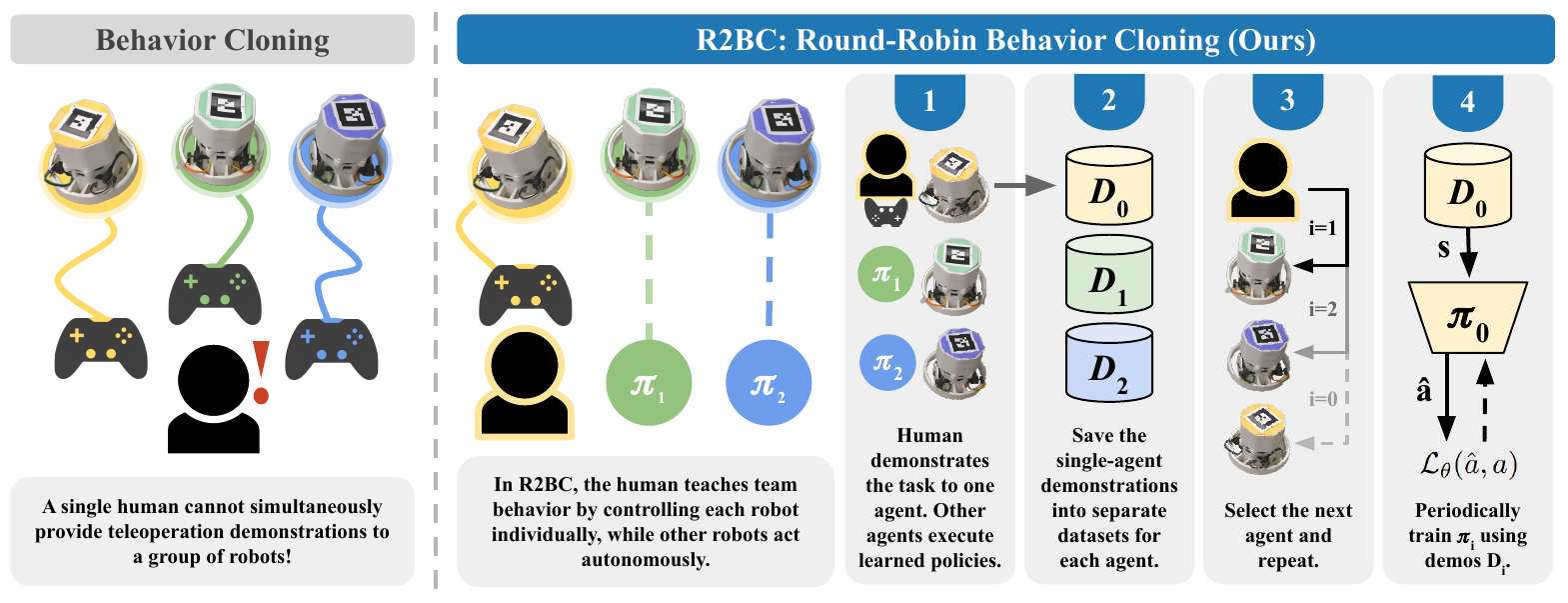}
\captionof{figure}{\textbf{Round-Robin Behavior Cloning (R2BC):}  Traditional Behavior Cloning (\textbf{left}) requires coordinated and centralized demonstrations, where an expert demonstrates actions near-optimally for all agents. In multi-agent domains, a single human cannot provide high-quality demonstrations due to underactuated control and increased cognitive load. Our method (\textbf{right}), R2BC, removes this restriction by letting the human control one agent at a time while the other agents execute their learned policies. This round-robin process collects realistic demonstrations and iteratively trains cooperative behavior. 
% Our method (\textbf{right}), R2BC, extends imitation learning to the multi-agent domain and only requires the human operator to provide demonstrations to one agent at a time. During each expert demonstration, the expert controls a specific agent while the other robots execute the learned policies cloned from their existing single-agent demonstrations, diversifying the distribution of observations seen by the expert's current agent and iteratively improving the agents' cooperative behavior.  
\label{fig:teaser}}
\end{strip}

%%%%%%%%%%%%%%%%%%%%%%%%%%%%%%%%%%%%%%%%
% Abstract
%%%%%%%%%%%%%%%%%%%%%%%%%%%%%%%%%%%%%%%%
\begin{abstract}
Imitation Learning (IL) is a natural way for humans to teach robots, particularly when high-quality demonstrations are easy to obtain. While IL has been widely applied to single-robot settings, relatively few studies have addressed the extension of these methods to multi-agent systems, especially in settings where a single human must provide demonstrations to a team of collaborating robots. In this paper, we introduce and study Round-Robin Behavior Cloning (R2BC), a method that enables a single human operator to effectively train multi-robot systems through sequential, single-agent demonstrations. Our approach allows the human to teleoperate one agent at a time and incrementally teach multi-agent behavior to the entire system, without requiring demonstrations in the joint multi-agent action space. We show that R2BC methods match—and in some cases surpass—the performance of an oracle behavior cloning approach trained on privileged synchronized demonstrations across four multi-agent simulated tasks. Finally, we deploy R2BC on two physical robot tasks trained using real human demonstrations. Videos, code, and supplemental materials can be found at \url{https://sites.google.com/view/r2bc/home}.
\end{abstract}

%%%%%%%%%%%%%%%%%%%%%%%%%%%%%%%%%%%%%%%%%%%%%%%
% INTRODUCTION
%%%%%%%%%%%%%%%%%%%%%%%%%%%%%%%%%%%%%%%%%%%%%%%
\section{Introduction}

Imitation Learning (IL) has become a cornerstone of robot learning, enabling robot agents to mimic human demonstrations without explicitly defined reward functions. 
In particular, multi-agent IL holds the potential to improve robot learning in a variety of applications involving multi-robot teaming and path planning, such as search and rescue~\cite{queralta2020collaborative}, disaster response~\cite{gregory2016application}, surveillance~\cite{scherer2020multi}, swarm robotics~\cite{li2019two}, and warehouse automation~\cite{bolu2021adaptive}. 
However, while single-agent IL has seen widespread success, the same cannot yet be said for multi-agent systems, where cooperation, coordination, and partial observability introduce unique challenges that are not easily addressed by simply scaling existing techniques. 
% As robotic platforms increasingly shift toward teams of agents working together to achieve shared goals, the question of how to effectively teach these systems using demonstrations remains both practically important and theoretically underexplored.

% Imitation Learning (IL) is a paradigm where a policy mimics the actions of a expert task demonstrator, enabling humans to teach robots how to accomplish manipulation~\cite{radosavovic2021state, von2024art, buamanee2024bi} and navigation~\cite{hussein2018deep} tasks though kinesthetic guidance or teleoperation. 
% Successful deployments of imitation learning policies on real-world robots include [TODO: More advancements here].
% Recent methods have shown that generalist cross-embodiment policies can be trained on a few task demonstrations and deployed in-the-wild on different robots~\cite{}, marking an important milestone for the field of robot learning.   

Several studies have examined cases where multi-agent IL could greatly improve solutions to real-world problems~\cite{andreychuk2025mapf, gao2025multi, bhattacharyya2018multi}. However, these works make strong assumptions about the type of demonstrations that can be provided to a group of agents, relying on coordinated and synchronous demonstrations in which all agents take correct actions simultaneously. Such synchronized demonstrations reduce the multi-agent IL problem to a single-agent problem in the joint action space of the agents. These methods are unrealistic, and usually implausible, for real world deployment, as human operators cannot reliably teleoperate multiple robots at the same time to accomplish complex tasks. 
% Another potential approach is Multi-Agent Reinforcement Learning (MARL)~\cite{marl-book}. MARL requires a well-defined reward, which are notoriously difficult to design correctly by hand~\cite{} and requires extensive training.

% Requiring the agents to work together often results in more nuanced definition of task success, which can be difficult to explicitly encode into a reward function. This reward design challenge appears to be a good indicator that behavior cloning or other LfD method may work well for multi-robot learning, but providing high-quality demonstrations to a fleet of robots can be challenging due to a deficit in actuation and control (teleoperation systems for MRS are often under-actuated).

In this paper, we address the following research question: \textit{How can we extend imitation learning to multi-agent systems when humans can only feasibly provide demonstrations to one agent at a time?}
Our work introduces a novel technique, \textbf{R}ound-\textbf{R}obin \textbf{B}ehavior \textbf{C}loning (\textbf{R2BC}) (Fig.~\ref{fig:teaser}), which enables a human demonstrator to provide single-agent demonstrations by cycling through the agents, while at any given time, all non-teleoperated agents %not currently receiving demonstrations 
execute their current learned policies. The key idea is that demonstrations can be individually provided to agents online to iteratively improve a multi-agent policy, achieving % that achieves 
parity with, or outperforming, the performance of traditional IL methods that require centralized demonstrations. The contributions of our work are as follows.

\begin{itemize}
    \item We introduce a novel problem setting, \textit{multi-agent imitation learning from single-agent demonstrations}, with the objective of training a collective policy using only individual agent demonstrations.
    \item We extend single-agent behavior cloning to this multi-agent problem setting by proposing Round-Robin Behavior Cloning (R2BC), which trains a policy that imitates the demonstrations while removing the unrealistic assumption in prior work of access to coordinated demonstrations.  
    % \item We provide theoretical analysis that shows that R2BC implicitly reduces the covariate shift at deployment when compared with traditional Behavior Cloning.
    \item We show that R2BC is able to match or exceed the performance of behavior cloning trained on joint action coordinated demonstrations in 4 simulated multi-agent domains using a synthetic demonstrator.
    \item We deploy R2BC in navigation and block pushing physical robot tasks using real human demonstrations and demonstrate that R2BC outperforms centralized BC by 3.25x and 5.9x on the two tasks, respectively. 
\end{itemize}
To the best of our knowledge, we are the first to propose and deploy a behavior cloning method for multi-agent systems that learns solely from online single-agent demonstrations.

%%%%%%%%%%%%%%%%%%%%%%%%%%%%%%%%%%%%%%%%%%%%%%%
% RW
%%%%%%%%%%%%%%%%%%%%%%%%%%%%%%%%%%%%%%%%%%%%%%%
\section{Related Work}
\subsection{Multi-Robot Teleoperation}
Providing demonstrations to a team of coordinating robots is a challenging research problem, as the number of degrees of freedom (DoF) required to control the entire system often exceeds what one human can control at once. Existing methods for teleoperating multi-agent systems include having a single human operator manually switch between  robots~\cite{farkhatdinov2008teleoperation}---e.g. at any given time, all but one robot receive no-op actions---or having a team of human operators who each control a different individual agent~\cite{patel2021transparency}.

% Learned methods include gesture-based teleoperation~\cite{de2022gestures},      

Recent studies have shown that learned models can utilize low-dimensional control inputs to control high-DoF systems such as swarms~\cite{turco2024reducing} and manipulators~\cite{losey2020controlling}. While these models can effectively allow humans to control complex systems (and therefore provide demonstrations), they are often trained on task-specific data and are unlikely to generalize to unseen tasks, making them a costly approach to providing demonstrations in deployed settings. Instead, we consider the problem setting in which a human can control a subset of the control parameters, corresponding to the control of an individual agent, for each demonstration.

Similarly, HiTAB (Hierarchical Training of Agent Behaviors)~\cite{squires2018lfd, luke2010learn} provides an approach to multi-agent demonstration collection that learns a set of atomic ``skills'' that a human can use as a discrete action set to control a group of heterogeneous agents. In this work, we assume that no preconceived skills library is available, and we provide demonstrations directly in the agent's action space. 

\subsection{Multi-Agent Imitation Learning (MAIL)}
The use of imitation learning (IL) to train a policy that mimics an expert task demonstrator has been widely studied in the single-agent literature~\cite{ross2011reduction, hoque2021thriftydagger, ross2010efficient, fang2019survey}. Extensions to multi-agent systems, namely \textit{Multi-Agent Imitation Learning (MAIL)}, have shown that policies can be learned from demonstrations to solve problems in grid energy management~\cite{gao2025multi}, autonomous vehicle control~\cite{bhattacharyya2018multi, huang2023multi}, and multi-agent path finding~\cite{andreychuk2025mapf}.

Current approaches successfully extend IL to the multi-agent setting, e.g. by modeling the spatial-temporal relationship between agents in a Graph Neural Network~\cite{zhou2019clone, li2025collective}, modeling latent structures of cooperation~\cite{le2017coordinated}, modeling individual agent deviations from an optimized coordinated policy~\cite{tang2024multi}, combining LfD with RL~\cite{huang2023multi, qi2024diffusion}, leveraging large transformers~\cite{andreychuk2025mapf}, and extending single-agent IL methods such as adversarial IL~\cite{bhattacharyya2018multi, song2018multi, fang2025multi} 
and inverse soft Q-learning~\cite{mai2024inverse} to the multi-agent regime.

Importantly, each of these proposed prior solutions involves training a policy from \textit{joint demonstrations}, where expert trajectories involve all $N$ agents working together simultaneously in noisily-optimal demonstrations. However, to obtain a dataset of joint demonstrations~\cite{andreychuk2025mapf}, one must assume unrealistic access to either synchronized human demonstrations for all agents or a pretrained RL demonstrator policy~\cite{zhou2019clone}, planner~\cite{li2019two, chen2023transformer, srinivasan2021fast}, or model for simulating expert behavior~\cite{zhou2019clone}, which may either remove the need to apply IL or 
%, in which case we have no need to deploy IL. These assumptions may 
be difficult to satisfy in real-world in-the-wild tasks.
Furthermore, none of the discussed prior methods directly study how to allow a real human to provide demonstrations to the agents.
Our work relaxes these assumptions, as a human is unlikely to be able to provide expert demonstrations to $N$ agents simultaneously, but is capable of providing examples to individual agents. 

Some work~\cite{hoque2023fleet} also considers methods for fleet imitation learning where robots are operating independently and multiple human supervisors are scheduled to correct several single-agent IL instances. Our work addresses scenarios where a single human must teach multiple robots that interact, observe, and coordinate with each other. We are the first to show successful multi-agent IL under realistic assumptions about the human's demonstration ability in multi-agent tasks. We show that a well-performing multi-agent policy can be trained by iteratively providing demonstrations to individual agents, one at a time, enabling training and deployment in settings with a single human supervisor.

%%%%%%%%%%%%%%%%%%%%%%%%%%%%%%%%%%%%%%%%%%%%%%%
% Problem
%%%%%%%%%%%%%%%%%%%%%%%%%%%%%%%%%%%%%%%%%%%%%%%
\section{Problem Formulation}
We seek to train a set of $N$ robot agents to collaborate to successfully accomplish a cooperative task. We formulate the problem of multi-agent learning from demonstrations as a Decentralized Partially Observable Markov Decision Process (Dec-POMDP) represented via the tuple $\left<\mathcal{I}, \mathcal{S}, \mathcal{A}, \mathcal{T}, \mathcal{R}, \Omega, \{\mathcal{O}_i\}_{i \in \mathcal{I}}, \gamma \right>$ consisting of a set of agents $\mathcal{I} = \{1, 2,\dots,N\}$, state space $\mathcal{S}$, and joint action space $\mathcal{A} = A_1 \times \dots \times A_N$. The environment transition dynamics, $\mathcal{T}:\mathcal{S} \times \mathcal{A} \times \mathcal{S} \to [0, 1]$, represent the probability of transitioning from state $s \in \mathcal{S}$ to state $s' \in \mathcal{S}$ after the agents take joint action $a \in \mathcal{A}$. The joint agent observation space $\Omega = \Omega_1 \times \ldots \times \Omega_N$ and corresponding observation function $\mathcal{O}_i: \mathcal{S} \times \mathcal{A} \times \Omega_i \rightarrow [0,1]$ determines the probability distribution over the possible observations received by agent $i$ after taking joint action $a$. As is standard in imitation learning, we do not learn directly from rewards ($\mathcal{R}$), but it is worth noting that we focus on games with \textit{shared reward} (also called \textit{common reward}), where all agents share the same collective objective.

We define a \textit{joint} expert trajectory $\tau$ with finite horizon $T$ as the sequence of states visited by an expert that controls all agents simultaneously, along with the associated joint expert actions taken in each state, $\tau = (s_0, a_0, s_1, a_1, \dots, s_T)$, where $s_t \in \mathcal{S}$ is the state at time $t$ and $a_t \in \mathcal{A}$ is the joint action (i.e. concatenation over $N$ individual agent actions) taken in state $s_t$.

Given a dataset of noisily-optimal demonstrations, $\mathcal{D}$, standard IL seeks to learn a joint policy $\pi:\mathcal{S}\to\mathcal{A}$ that closely models the expert's state-action pairs and achieves similar task performance. Generally, the goal of IL is to find the parameters $\theta$ to a policy $\pi_{\theta}$ that minimize the error between the demonstrated expert's actions, $\pi^*(s)$, and the outputs of the learned policy, $\pi_\theta(s)$, expressed via the discrepancy loss:

\begin{equation}
    \min_\theta \mathbb{E}_{\tau \sim \mathcal{D}} \left[ \sum^{T-1}_{t=0}\mathcal{L}(\pi_\theta(s_t), \pi^*(s_t)) \right],
    \label{eq:traditional-bc}
\end{equation}
where $\mathcal{L}$ is a loss function that penalizes divergence from the expert's actions, commonly the squared error.
% $||\pi_\theta(s_t) -\pi^*(s_t)||^2$.

Contrary to the single-agent IL literature, we assume that a single human \textit{cannot} teleoperate the entire system with optimal actions, and therefore cannot provide actions in the joint action space, $\mathcal{A}$. For systems with any non-arbitrary number of agents (i.e. $N \geq 2$), this is a reasonable assumption given that (1) the amount of cognitive burden required to monitor the system increases with the number of agents~\cite{kaduk2024one, turco2024reducing} and (2) teleoperation interfaces cannot control all degrees of freedom for complex systems without custom control schemes~\cite{turco2024reducing, aggravi2021decentralized}. Instead, we assume that the human is only capable of controlling small parts of the system, such as one individual agent at a time, which we call \textit{single-agent demonstrations}. For agent $i \in \mathcal{I}$, a demonstration of length $T$ takes the form
\begin{equation}
    \tau_i = (o_{i, 0}, a_{i, 0}, o_{i, 1}, a_{i, 1}, \dots, o_{i, T})
\end{equation}
where $o_{i, t} \in \Omega_i$ is the $i$\textsuperscript{th} agent's local observation, and $a_{i,t} \in A_i$ is agent $i$'s \textit{individual action}, rather than the joint action of all agents. 
% Note that the state remains unchanged here, and the other agents will still be operating in the environment, as should be reflected in the expert actions.
We seek to train multi-agent policies that achieve sufficient collective performance using only $N$ sets of single-agent demonstrations, $\{\mathcal{D}_1, \mathcal{D}_2, \dots, \mathcal{D}_N\}$, where $\mathcal{D}_i$ is the set of single-agent demonstrations for agent $i$. 

% As we describe in the subsequent section, our method utilizes a novel online, on-policy approach that reduces the covariate shift between the training data and the state seen in deployment.

%%%%%%%%%%%%%%%%%%%%%%%%%%%%%%%%%%%%%%%%%%%%%%%
% Methods
%%%%%%%%%%%%%%%%%%%%%%%%%%%%%%%%%%%%%%%%%%%%%%%
\begin{figure*}
    \centering
    \includegraphics[width=1\linewidth]{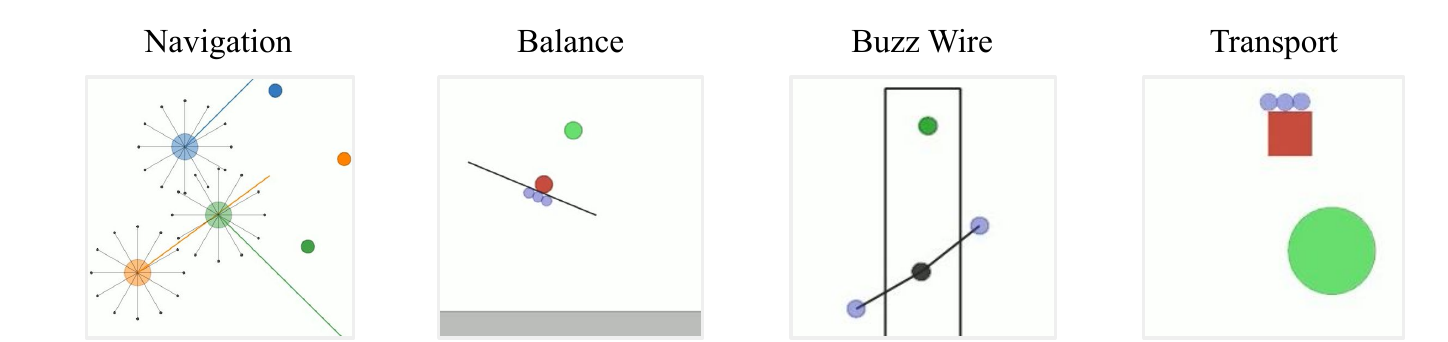}
    \includegraphics[width=1\linewidth]{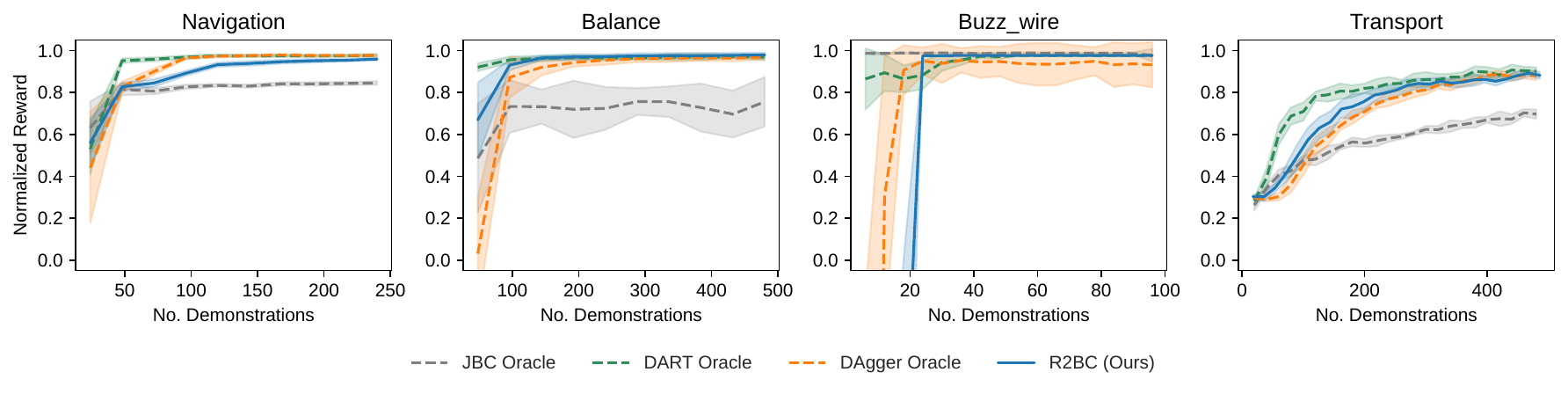}
    \caption{\textbf{Simulation Results:} We compare the R2BC method to a set of three baselines that assume oracle (privileged) access to a joint action demonstrator.  
    % behavior cloning on the joint action space (JBC), which assumes full coordinated demonstrations for $N$ agents.
    Results are averaged over 10 seeds with shaded regions indicating a 99\% confidence interval. 
    \textbf{Top:} Four multi-agent tasks selected from the Vectorized Multi-Agent Simulator (VMAS)~\cite{bettini2022vmas}. 
    \textbf{Bottom:} The performance of each method is shown with increasing number of expert demonstrations. In each case, R2BC outperforms or achieves parity with all baseline oracle methods, despite not having privileged access to joint action demonstrations. Performance is normalized between 0.0 (random policy) and 1.0 (oracle demonstrator performance). 
    }
    \label{fig:res-main}
\end{figure*}
\section{Round-Robin Behavior Cloning}
\label{sec:methods}
% In this section, we introduce Round-Robin Behavior Cloning (R2BC) (Fig.~\ref{fig:teaser}), a method that enables a single human operator to cycle through each agent individually when providing task demonstrations. 
% To iteratively learn a collaborative policy, the demonstrator continues to provide actions (in a round-robin fashion) for each agent while each remaining agent executes its current learned policy.

\begin{algorithm}[t]
\caption{Round-Robin Behavior Cloning (R2BC)}
\label{alg:R2BC}
\begin{algorithmic}[1]
\Require Number of agents $N$, expert policy $\pi^*$, initial agent policies $\{\pi_1, \dots, \pi_N\}$, demonstration buffers $\{
\mathcal{D}_1, \dots, \mathcal{D}_N\}$, update frequency $k$, and task horizon $T$.
\State Initialize iteration counter $c \gets 0$
\While{c $\leq$ MAX\_ITERS}
    \For{each agent $i = 1$ to $N$} \Comment{round-robin}
        \State Reset environment to initial state $s_0 \sim \mathcal{S}_0$
        \For{$t = 0$ to $T$}
            \State Agent $i$ receives expert action $a_{i,t} \gets \pi^*(o_{i, t})$
            \For{each agent $j \ne i$}
                \State Agent $j$ computes action $a_{j,t} \gets \pi_j(o_{j, t})$
            \EndFor
            \State Execute joint action $a_t = (a_{1,t}, \dots, a_{N,t})$
            \State Receive next observation $o_{i, t + 1}$
            \State Append $(o_{i, t}, a_{i,t})$ to $\mathcal{D}_i$
        \EndFor
    \EndFor
    \If{$c \bmod k = 0$}
        \State Update $\pi_{i}$ using BC on $\mathcal{D}_i \ \forall i\in \mathcal{I}$.
    \EndIf
    \State $c \gets c + 1$
\EndWhile
\end{algorithmic}
\end{algorithm}
% \subsection{Round-Robin Behavior Cloning}
\label{sec:R2BC-details}
To teach cooperative tasks to teams of robots using single-agent demonstrations, we propose \textbf{Round-Robin Behavior Cloning (R2BC)}, an online imitation learning algorithm that enables a single human operator to sequentially provide demonstrations to a team of cooperative agents. Unlike traditional joint action behavior cloning, which requires the expert to simultaneously demonstrate actions for all agents (a setting infeasible for real-world human teleoperation), R2BC assumes that the expert can only control a single-agent at any given time. Our method, shown in Algorithm~\ref{alg:R2BC},  cycles through the agents in a round-robin fashion, allowing the expert to demonstrate behaviors for one agent while the others operate their current learned policies. After an episode of demonstrating the task to agent $i$, the demonstrator switches to agent $i + 1 \ (mod \ N)$. Providing round-robin demonstrations ensures a uniform spread of demonstrations across all $N$ agents during training.
% This setup enables the realistic collection of diverse, on-policy training data across a wide distribution of states.

In a decentralized fashion, each agent maintains its own policy $\pi_i : \Omega_i \rightarrow \mathcal{A}_i$. Each agent $i$ stores its demonstrations within a buffer $\mathcal{D}_i$, which is used only to train its policy, $\pi_{\theta_i}$. At regular intervals (every $k$ demonstrations), each agent's policy is updated using that agent's collected demonstration dataset using the standard behavior cloning loss:
\[
L(\theta_i) = \sum_{(o_i, a_i) \sim \mathcal{D}_i} \| \pi_{\theta_i}(o_i) - a_i \|^2_2.
\]

R2BC combines two extensions to traditional centralized behavior cloning. First, R2BC allows for full decentralization, both in policy execution and demonstrations. The main motivation of our work is to address realistic demonstrator limitations by decentralizing demonstrations, allowing iterations of single-agent demonstration to teach behavior to multi-robot systems. Policy decentralization also addresses several real-world deployment considerations such as limited communication and partial observability. The full decentralization of R2BC promotes realistic demonstration collection from real humans and deployment on physical hardware in the real world.

Second, R2BC iteratively improves team performance in an online-learning manner, where demonstrations are provided online while other agents execute policies cloned on the demonstrations observed so far. Our method directly benefits from an online regime because demonstrators must teach a single-agent to perform appropriately alongside \textit{existing actors}, i.e. the other agents that are executing autonomous policies. This ideally allows a demonstrator to teach an agent how to act amidst wide variability in the help received from other agents, resulting in greater demonstration diversity and downstream resilience to error and distribution shift. In Section~\ref{sec:ablations}, we examine several ablations of R2BC and highlight that both extensions---decentralization and online policy learning---are essential to the success of our method.

% \subsection{Reduced Covariate Shift}\label{sec:shift-theory}
% Compared to behavior cloning, we hypothesize that R2BC implicitly reduces the covariate shift by diversifying the behavior of the other agents while demonstrating optimal actions to the $i$-th agent. A centralized behavior cloning paradigm would require coordinated near-optimal actions jointly taken by all agents, limiting the training states to only states where all agents are acting optimally. R2BC relaxes this distribution and allows states where only the demonstrating agent is acting optimally.

% While we do not provide a theoretical foundation for reduced covariate shift in this work, we provide empirical evidence for this claim in our experiments and plan to prove this in future work. We believe our method may have similar theoretical improvements as other online imitation learning methods such as DAGGER~\cite{ross2011reduction} and demonstrator noise injection methods like DART~\cite{laskey2017dart}. 

%%%%%%%%%%%%%%%%%%%%%%%%%%%%%%%%%%%%%%%%%%%%%%%
% SIMULATED EXPERIMENTS
%%%%%%%%%%%%%%%%%%%%%%%%%%%%%%%%%%%%%%%%%%%%%%%
\section{Simulated Experiments} \label{sec:sim-exp}
We compare R2BC to three oracle imitation learning methods on four cooperative tasks in simulation, each requiring coordination between 2-3 robots. First, we introduce the simulated scenarios and experimental details (Section~\ref{sec:sim-exp-design}). Second, we highlight the surprising performance of R2BC compared to a privileged Joint Behavior Cloning (JBC) baseline, where demonstrations are centralized, synchronous, and near-optimal (Section~\ref{sec:sim-jbc}). Finally, we offer intuition about the online benefits of R2BC by comparing our method to oracle implementations of DAgger~\cite{ross2011reduction} and DART~\cite{laskey2017dart} trained on online joint demonstrations (Section~\ref{sec:sim-online-il}).

\subsection{Experiment Design}\label{sec:sim-exp-design}
We evaluate R2BC and three baselines on four simulated cooperative tasks using the Vectorized Multi-Agent Simulator (VMAS)~\cite{bettini2022vmas}, shown in Figure~\ref{fig:res-main} (top).

\textbf{Navigation}: Three agents are randomly positioned in a two-dimensional space each with a designated goal. Agents can utilize LiDAR-style sensors to sense other agents and avoid collisions \textit{en route} to their objectives.

\textbf{Balance}: Three agents are placed under a freely rotating line carrying a spherical package. The agents must transport this package  from the bottom to a goal at the top, without allowing the line or the package to fall.

\textbf{Buzz Wire}: Two agents are connected to an enclosed mass using rigid linkages. The agents are penalized for coming in contact with the enclosure and must take cooperative actions that push the mass through the hallway. This environment represents a system with coupled dynamics, where the actions of one agent can directly displace the other agent, reflecting difficult multi-agent transition dynamics.

\textbf{Transport}: Three agents cooperate to push a package into a designated goal region. The packages are heavy and cannot be pushed by individual agents alone, requiring agents to work together to transport the package.

%For each environment, 
The methods are compared under a demonstration budget that varies between domains depending on the task difficulty. Each policy is trained using a MSE-Loss until the loss plateaus, using a fixed learning rate of $0.001$ and a minibatch size of 256 state-action pairs. After training, we evaluate the learned policies on an identical set of 50 environment initialization seeds and compute the average environment reward. In all experiments, we normalize the task reward to the range $[0, 1]$, where a score of $0.0$ corresponds to a random policy and $1.0$ reflects the performance of the oracle expert demonstrator. As is standard in IL, the expert defines an upper bound that is typically not exceeded by the learner. Additional procedures, hyperparameters, and experiments can be found %in our supplemental materials 
on our supplemental website\footnote{https://sites.google.com/view/r2bc/home}.

\begin{figure}
    \centering
    \includegraphics[width=0.9\linewidth]{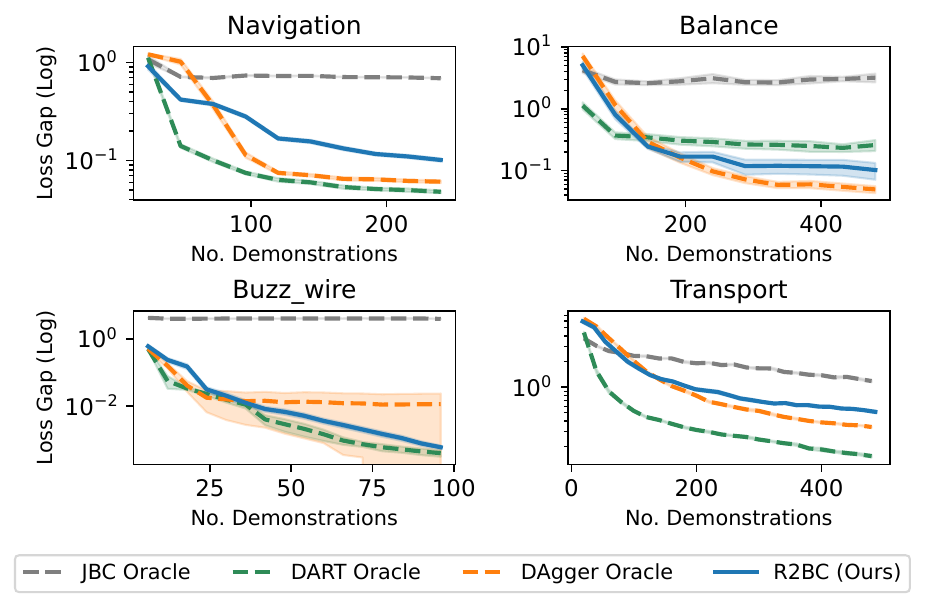}
    \caption{\textbf{Covariate Shift Reduction:} The \textit{Loss Gap} is defined as the difference between the training loss on the demonstration set and testing loss on evaluation trajectories (lower is better). Dashed lines indicate methods that require privileged centralized demonstrations, while the solid blue line is our method. R2BC reduces the train–test loss gap compared to JBC, while showing benefits comparable to other online imitation learning methods. Shaded regions indicate standard error across 10 seeds.
    }
    \label{fig:covariate-shift}
\end{figure}

\subsection{Comparison to Joint Action Behavior Cloning (JBC)} \label{sec:sim-jbc}
JBC serves as an oracle behavior cloning method that learns from coordinated demonstrations provided in the joint action space. %Recall from our problem statement that 
In practice, such demonstrations require complex learned teleoperation schemes or many-to-many teleoperation schemes where $N$ human operators control $N$ robots. Our experiments utilize a multi-agent reinforcement learning policy (MAPPO~\cite{yu2022surprising, schulman2017proximal}) trained on the ground truth environment reward to provide coordinated demonstrations to JBC for supervised imitation learning. Therefore, we consider this an oracle baseline with access to unrealistic data that reflects prior assumptions about coordinated demonstrations in multi-agent IL~\cite{andreychuk2025mapf, gao2025multi, bhattacharyya2018multi}.

While it might seem that JBC would be an upper-bound for R2BC's performance---due to JBC having oracle access to near-optimal, coordinated demonstrations---surprisingly, we found that in all simulated domains, R2BC outperforms or achieves parity with the oracle JBC method (Figure~\ref{fig:res-main}). Although JBC learns from coordinated demonstrations, its demonstration data solely contains examples of near-optimal behavior and lacks examples of how to correct erroneous behaviors or act in out-of-distribution states. Conversely, R2BC benefits from observation perturbations caused by the non-demonstrating agents, which act autonomously. The initially unpredictable actions of the other agents create situations where the single-agent demonstrator can show the robot how to correct wayward cooperative behavior or avoid a collision with another agent. Cases like these 
are much less prevalent in JBC demonstrations, in which the expert demonstrates fully-coordinated agent behaviors.
%would be absent from JBC demonstrations, which assume near-optimal coordination where agents only observe good behavior from their teammates. 
Our empirical observations offer support for R2BC as a realistic and viable alternative to JBC, in several cases surpassing the performance of training on ``super-human'' JBC demonstrations using only single-agent demonstrations.

% In \textit{navigation}, \textit{balance}, and \textit{buzz wire}, R2BC achieves better performance than JBC after 96, 144, and 54 single-agent demonstrations, respectively. These results indicate strong support for H1 in the context of independent policies, with some support for H1 in the centralized case. In each environment, R2BC reliably achieves parity with, or surpasses, JBC---despite JBC having access to unrealistic, fully coordinated joint action demonstrations. 
% Ablation Figure
\begin{figure}
    \centering
    \includegraphics[width=1\linewidth]{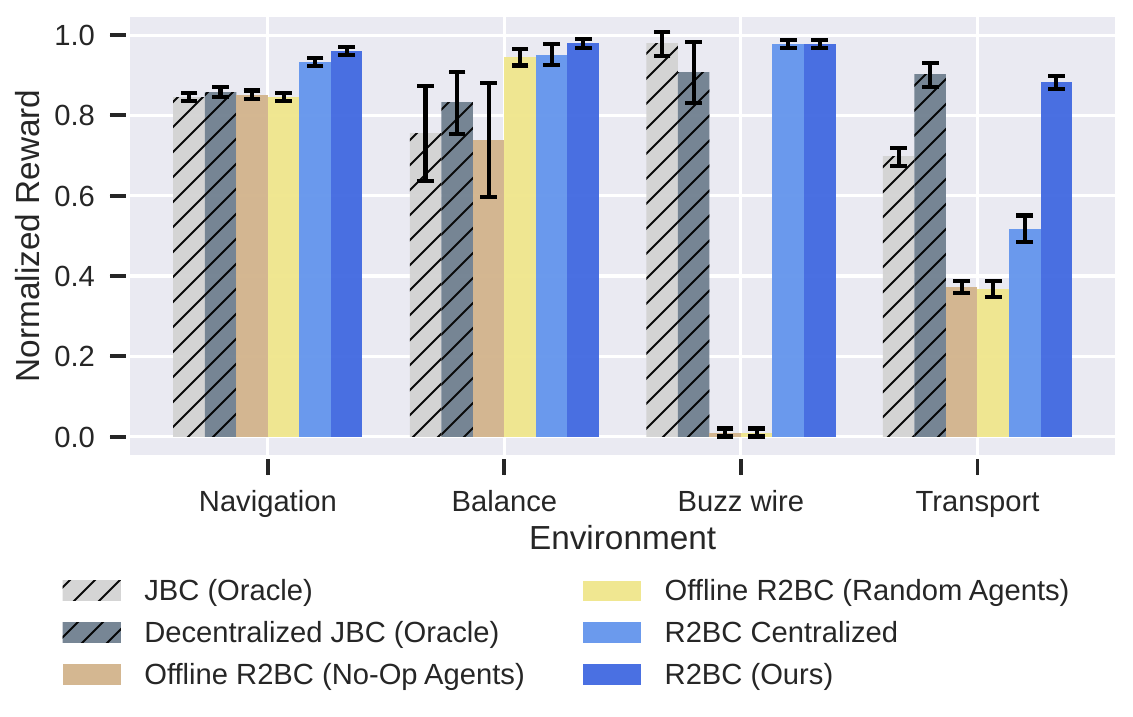}
    \caption{\textbf{R2BC Ablations}: Comparison of R2BC to several ablations across four simulated environments. Across all environments, R2BC consistently outperforms the two offline variants and matches or exceeds the oracle JBC methods. Error bars indicate 99\% confidence interval over 10 seeded runs.}
    \label{fig:ablations}
\end{figure}

\subsection{Comparison to Online Imitation Learning Baselines} \label{sec:sim-online-il}

We hypothesize that R2BC's significant performance improvement over JBC results from the online paradigm, which exposes agents to diverse states and levels of cooperation from teammates. We compare R2BC to two popular single-agent online imitation learning methods specifically designed to combat compounding errors:  (1) DAgger~\cite{ross2011reduction}, where a joint action expert provides corrective labels to on-policy rollouts 
% The supervisors labels are used to update the policy before collecting more online corrections.
and (2) DART~\cite{laskey2017dart}, where noise is injected into the demonstrator's teleoperation online to vary the state distribution and allow for online corrections.
% DART~\cite{laskey2017dart} and DAgger~\cite{ross2011reduction}, which offer an online IL comparison for a joint action oracle demonstrator. 
Our implementation of these methods in multi-agent domains uses oracle centralized corrections generated from the same multi-agent RL policy described in the previous section.

Our simulated results for DART and DAgger indicate that R2BC achieves similar performance to both privileged online baselines (Figure~\ref{fig:res-main}). %In every environment, R2BC exhibits similar benefits to DART and DAgger where 
In all three cases, increasing the number of online demonstrations correlates with improved policy performance. These results provide evidence that R2BC yields a similar reduction in compounding errors to DART and DAgger, but importantly, R2BC achieves this without requiring $N$ human demonstrators or a super-human joint demonstrator. We hypothesize that the observation variance induced by the non-demonstrating agents in R2BC acts as artificial noise injected into the demonstrator's state, serving a similar function to the noise injected in DART, which was shown to reduce covariate shift in single-agent IL.
%and draw a similar conclusion. 
We provide empirical support for covariate shift reduction in Figure~\ref{fig:covariate-shift}, which shows that R2BC exhibits similar trends in the train-test loss gap as the two oracle online baselines, while achieving much better generalization than JBC.

% Another finding from Figure~\ref{fig:res-main} is that the performance of oracle methods is almost always better than our methods for a very low number of demonstrations. Intuitively, this can be explained by the online nature of the R2 methods which start from scratch to iteratively achieve coordination, rather than having initial access to a coordinated demonstrator. Notably, the performance of these oracle methods ultimately plateau and lack the same performance as their online round-robin counterparts, which continue to improve many demonstrations after JBC has converged.

% Real Robot Figure
\begin{figure}
    \centering
    \includegraphics[width=1\linewidth]{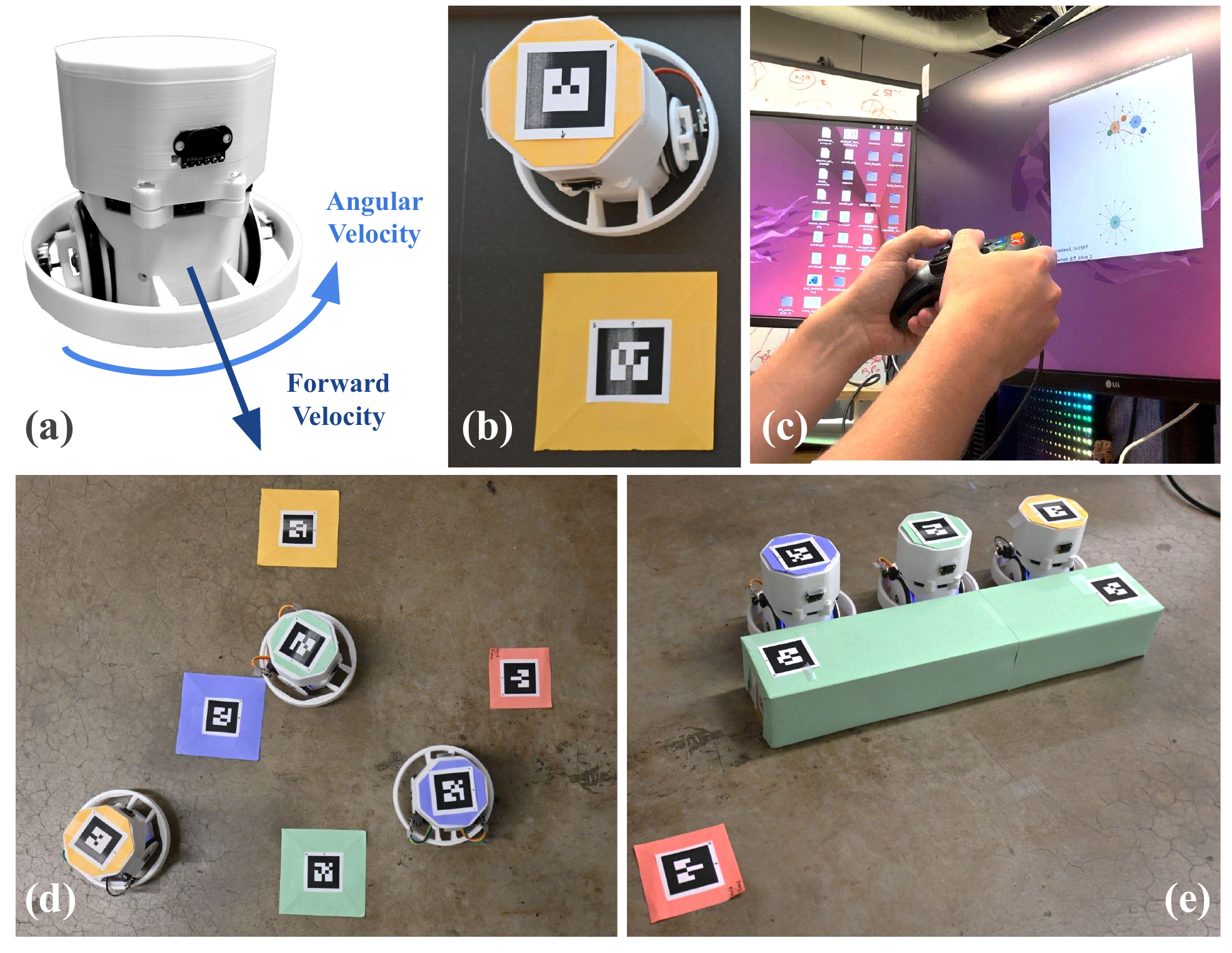}
    \caption{\textbf{Real Robot Experiments:} \textbf{(a)} The HeRo+ Robot, a differential drive mobile robot that acts using a forward velocity and angular velocity, \textbf{(b)} AruCo markers used to track robot/goal states, \textbf{(c)} R2BC demonstrations are provided in simulation by a real human, \textbf{(d)} a real world instantiation of the navigation task (from Figure~\ref{fig:res-main} top), where agents must navigate to their colored goals while avoiding collisions, \textbf{(e)} a block-pusher task where robots must transport a 0.5kg box across the environment. In both (d) and (e), the pink AruCo marker is used for camera-to-world registration.}
    \label{fig:real-world}
\end{figure}

%%%%%%%%%%%%%%%%%%%%%%%%%%%%%%%%%%%%%%%%%%%%%%%
% ABLATIONS
%%%%%%%%%%%%%%%%%%%%%%%%%%%%%%%%%%%%%%%%%%%%%%%
\section{Simulated Ablations}
\label{sec:ablations}

In Section~\ref{sec:methods}, we highlighted the two extensions to traditional centralized behavior cloning introduced by our method: full decentralization and online learning. In this section, we perform several ablation experiments and show that both of these elements (decentralization and online learning) are essential to the training and deployment of our method. We evaluate the performance of these ablations and summarize the findings in Figure~\ref{fig:ablations}. All experiments shown use simulated human feedback.

\textbf{Joint Behavior Cloning (JBC)} most closely reflects traditional behavior cloning, where a policy maps states to the \textit{joint actions} for all agents in a centralized manner. We compare this with \textbf{Decentralized JBC}, an approach that utilizes the same offline oracle joint action demonstrations as JBC, but clones actions into $N$ independent networks. Notably, this method also decentralizes the state information, and the individual agents act solely based on their local (potentially partially observable) state. We find that Decentralized JBC performs on par with JBC in the navigation and buzz wire tasks, and significantly better than JBC in the balance and transport tasks. However, this oracle method still requires centralized near-optimal demonstrations, which a lone human demonstrator is unlikely to be able to provide.

Next, we evaluate two decentralized versions of R2BC that are offline (\textbf{Offline R2BC}). In these methods, the human cycles round-robin through each agent and demonstrates the task but, unlike in R2BC, the agents do not regularly update their policies and therefore do not learn during the demonstration phase. We test two versions of this: in the first, the non-demonstrating agents do not execute any actions (\textbf{No-Op Agents}), while in the second, the non-demonstrating agents uniformly sample random actions from the action space (\textbf{Random Agents}). We find that these approaches can occasionally perform on par with R2BC (e.g., the Random variant in Balance), but overall they are unreliable and prone to high variance across tasks, as seen in Buzz Wire.

Finally, we consider a centralized version of R2BC (\textbf{Centralized R2BC}). Here, instead of learning $N$ decentralized policies, we maintain a single centralized network that outputs the joint action, and during training, we mask the loss to only update the output layer neurons corresponding to the demonstrated agent. We find that Centralized R2BC performs slightly worse than the decentralized version in navigation and balance, comparably in buzz wire, and significantly worse in transport.

These ablations highlight the importance of both decentralization and online learning during demonstration collection: removing either element leads to unreliable or degraded performance across tasks. Together, these findings demonstrate that the full R2BC framework is necessary to achieve robust multi-agent learning from single-agent demonstrations.

%%%%%%%%%%%%%%%%%%%%%%%%%%%%%%%%%%%%%%%%%%%%%%%
% REAL ROBOTS
%%%%%%%%%%%%%%%%%%%%%%%%%%%%%%%%%%%%%%%%%%%%%%%
\section{Physical Robot Deployment}

We deploy both JBC and R2BC on three HeRo+ robots~\cite{mattson2025discovery, rezeck2023hero}, each with a two-dimensional action space consisting of forward/backward velocity and angular turning velocity (Figure~\ref{fig:real-world}a). A centralized Robot Operating System (ROS) server computes the instantaneous state of the environment using a Realsense D435i camera to track AruCo markers on agents and objects (Figure~\ref{fig:real-world}b), and passes either local observations to each agent’s policy (R2BC) or the full state to a centralized policy (JBC). Our real world experiments consist of 2 tasks:

\textit{Navigation} (Figure~\ref{fig:real-world}d): A physical version of the simulated navigation task, as described in section~\ref{sec:sim-exp-design}. Mock LiDAR values are computed instantaneously using the centralized video feed.

\textit{Block Pusher} (Figure~\ref{fig:real-world}e): A physical variation of the simulated transport task. Three agents must coordinate to push an elongated green block to a goal location. Because of the block’s shape, agents must balance forces applied at different contact points to successfully transport the block.

% \subsection{Human Demonstration Collection}
To evaluate JBC and R2BC using real human demonstrations, the authors provided two sets of 240 demonstrations for each task by teleoperating agents in simulation using an Xbox360 joystick controller (Figure~\ref{fig:real-world}c). In R2BC, the demonstrator provided single-agent demonstrations and they were aware of which agent they would be controlling prior to the start of each episode. In JBC, the demonstrator used two game controllers and could control all agents simultaneously using 3 of the analog sticks. In practice, this often meant focusing on only 1–2 agents at once, or attempting to span the inputs by placing the controllers on a flat surface and manipulating the analog sticks with both hands, illustrating the difficulty of providing joint demonstrations in real time.

To support sim2real transfer, the simulated environments for both navigation and block pushing were customized to match the kinematics of our HeRo+ robots and physical workspace. Similarly, the action space in the simulator was clamped to realistic linear and angular velocities achievable by our robots in the real world. 
% The subsequent section highlights the results of direct sim2real deployment, where the policies trained with JBC and R2BC in simulation are zero-shot deployed and evaluated in the real world.

% \subsection{Results}

\begin{figure}
    \centering
    \includegraphics[width=0.95\linewidth]{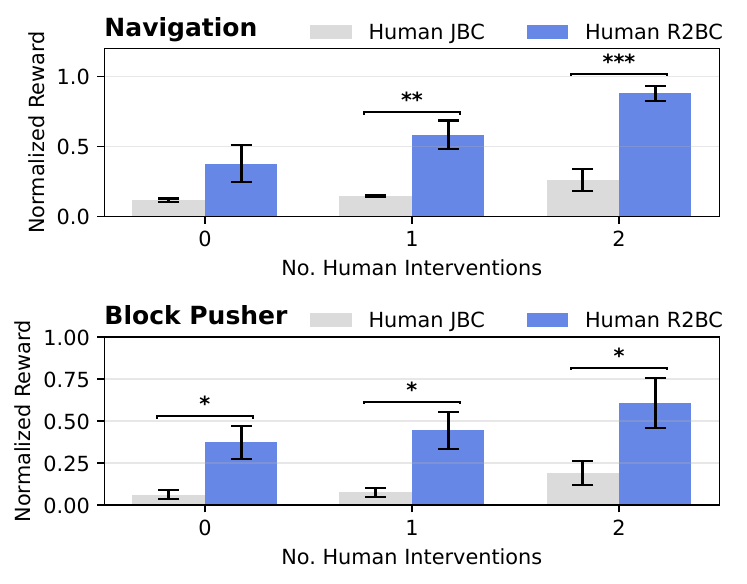}
    \caption{\textbf{Real Robot Results:} JBC and R2BC policies trained from real human demonstrations in simulation are deployed directly onto the real robots in two tasks.
    To account for sim-to-real distribution shift, we report results with 0, 1, and 2 online interventions, where a human supervisor can briefly correct robot behavior. All results show mean cumulative episode reward with standard error over 5 randomized initial configurations. Stars (*) indicate one-tailed significance: *$p < 0.05$, **$p < 0.01$, ***$p < 0.005$.
    % To offer a fair comparison of methods and account for sim2real distribution shift, we show results for deployments with 0, 1, and 2 online interventions, where a human supervisor can correct a robot's behavior using remote teleoperation. All results display the mean cumulative episode reward and standard error over 5 trials with different initial environment configurations. Stars (*) indicate one-tailed statistical significance *$p < 0.05$, **$p < 0.01$, ***$p < 0.005$.
    }
    \label{fig:real-results}
\end{figure}
For these physical tasks, we deployed the trained policies directly onto the real robots and evaluated the quantitative task performance over a fixed horizon. For the navigation task, we measured the cumulative inverse distance between agents and their respective goals (same reward metric used to evaluate policies in simulation) with a task horizon of 90 seconds. For the block pushing task, we measured the cumulative distance the package moved in the direction of the goal region with a task horizon of 30 seconds. Due to the distribution shift that accompanies sim2real deployment, deployed behaviors can sometimes get stuck in local minima, and agents are unable to complete the task. We account for this potential distribution shift in our real-world experiments by studying the effect of allocating 0, 1, or 2 human interventions. In each intervention, the human supervisor selects a robot that appears ``stuck'' and manually controls its actions for 3 seconds to correct the robot's behavior while the other robots continue to execute their autonomous policies.

% Any sub-optimal behavior that occurs after all allocated interventions have been used is uncorrected for the remainder of the task.

Our results (Figure~\ref{fig:real-results}) are averaged over 5 initial environment configurations, which were randomized but consistent across comparisons. To test for the directional hypothesis that R2BC performs significantly better than JBC, we employ a Welch two-sample t-test and convert the two-tailed p-value to a single-tail p-value.
% Videos of the real-robot experiments can be found at our website.

In the navigation task, we find that R2BC performs 3.25x better than JBC on average without supervisor interventions; however, the statistical tests do not indicate significance ($p=0.0773$) due to the high performance variance that both methods experience under zero-shot sim2real deployment. However, significant differences between R2BC and JBC emerge when allowing both methods access to supervisor interventions. With one intervention allowed, R2BC outperforms JBC by 3.92x ($p=0.0094$) and with two interventions allowed, R2BC outperforms JBC by 3.36x ($p<0.005$). In the block pushing task, we find that R2BC performs 5.9x better than JBC ($p=0.0126$) with no supervisor interventions. With one intervention allowed, R2BC outperforms JBC by 5.84x ($p=0.0107$) and with two interventions allowed, R2BC outperforms JBC by 3.16x ($p=0.0239$).

Ultimately, we find that training with real human demonstrations leads to improved performance for R2BC compared to JBC. As expected in sim2real transfer, the deployed policies encounter some local minima that can be easily corrected with just a few seconds of supervisor intervention. Notably, supervisor corrections help the performance of both methods, yet slight corrections to R2BC policies appear to have a much greater effect on performance compared to equivalent corrections in JBC. This reinforces the notion that R2BC is a learning paradigm that leads to high performing policies and can easily be adopted by teachers, removing the need to collect joint action demonstrations. Our results confirm that Round-Robin Behavior Cloning can reliably outperform traditional centralized behavior cloning methods both in simulation and when trained with real human demonstrations and deployed on physical hardware.

%%%%%%%%%%%%%%%%%%%%%%%%%%%%%%%%%%%%%%%%%%%%%%%
% CONCLUSION
%%%%%%%%%%%%%%%%%%%%%%%%%%%%%%%%%%%%%%%%%%%%%%%
\section{Conclusion \& Future Work} 
\label{sec:conclusion}

Our paper highlights a novel research question targeting the deployment of multi-agent systems: \textit{How can we extend imitation learning to multi-agent systems when humans can only feasibly provide demonstrations to one agent at a time?} While multi-agent IL methods assume unrealistic joint demonstrations, our algorithm, R2BC, utilizes round-robin single-agent demonstrations to gather examples of what each agent should do with respect to the environment and other agents. Our results show that R2BC can outperform JBC in both simulated domains with synthetic demonstrations and physical robot deployment with real human demonstrations.

One underexplored aspect of this work is evaluating the human factors associated with R2BC. In future work, we plan to conduct user studies with non-experts to evaluate the cognitive burden and qualitative experience of our method. 
% Additional analysis will also help to determine the domains and tasks for which R2BC is best suited, as our experiments cover only a subset of multi-agent scenarios. 
Although the empirical evidence for R2BC is strong, developing theoretical guarantees for convergence and covariate shift reduction would further reinforce the intuition provided in this work. Overall, R2BC provides a practical path toward teaching multi-agent systems via realistic human demonstrations, bridging the gap between single-agent IL and scalable multi-agent deployment.

% Although the empirical evidence for R2BC is strong, theoretical guarantees for convergence and covariate shift reduction will help support the intuition provided in this work.

% How to take this offline? (Safe for real-world testing?) Synthetic augmentations to achieve the same outcome? 

% Using these online demonstrations, we show that a synthetic demonstrator can actually achieve greater task performance using our algorithm than when providing joint action coordinated demonstrations for all $N$ agents simultaneously. We hypothesize that the online nature of R2BC is able to implicitly reduce the covariate shift compared to JBC and therefore correctly infer actions in unseen states at runtime.

% We are eager to continue this work and enable the real-world deployment of R2BC. While our initial results are promising, our results indicate that the reliability of our method may depend on the type of task/environment that R2BC is deployed in. Therefore, we plan to conduct additional analysis to determine which domains and tasks R2BC is best suited for, including testing on a more diverse set of environments. We also plan to deploy this method for real-world mobile robotics tasks and evaluate the cognitive burden on humans providing R2BC feedback online to robots in the real world.

\section*{Acknowledgments}
This work was completed in the Aligned, Robust, and Interactive Autonomy (ARIA) Lab at the University of Utah. ARIA Lab research is supported in part by the NSF (IIS-2310759, IIS2416761), the NIH (R21EB035378), ARPA-H, Coefficient Giving, and the ARL STRONG program.

\bibliographystyle{plain}
\bibliography{references}

@inproceedings{zhou2019clone,
  title={Clone swarms: Learning to predict and control multi-robot systems by imitation},
  author={Zhou, Siyu and Phielipp, Mariano J and Sefair, Jorge A and Walker, Sara I and Amor, Heni Ben},
  booktitle={2019 IEEE/RSJ International Conference on Intelligent Robots and Systems (IROS)},
  pages={},
  year={2019},
  organization={IEEE}
}

@inproceedings{qi2024diffusion,
  title={Diffusion-Based Multi-Agent Reinforcement Learning with Communication},
  author={Qi, Xinyue and Tang, Jianhang and Jin, Jiangming and Zhang, Yang},
  booktitle={Asia Pacific Wireless Communications Symposium},
  pages={},
  year={2024}
}

@article{li2025collective,
  title={Collective Behavior Clone with Visual Attention via Neural Interaction Graph Prediction},
  author={Li, Kai and Ma, Zhao and Li, Liang and Zhao, Shiyu},
  journal={arXiv preprint arXiv:2503.06869},
  year={2025}
}

@inproceedings{le2017coordinated,
  title={Coordinated multi-agent imitation learning},
  author={Le, Hoang M and Yue, Yisong and Carr, Peter and Lucey, Patrick},
  booktitle={International Conference on Machine Learning},
  pages={},
  year={2017},
  organization={PMLR}
}

@article{tang2024multi,
  title={Multi-Agent Imitation Learning: Value is Easy, Regret is Hard},
  author={Tang, Jingwu and Swamy, Gokul and Fang, Fei and Wu, Steven Z},
  journal={Advances in Neural Information Processing Systems},
  volume={},
  pages={},
  year={2024}
}

@article{gao2025multi,
  title={Multi-Agent Imitation Learning Based Energy Management of a Microgrid With Hybrid Energy Storage and Real-Time Pricing},
  author={Gao, Shuhua and Xu, Yizhuo and Zhang, Zhaoqian and Wang, Zhengfang and Zhou, Xiaoyu and Wang, Jing},
  journal={IEEE Internet of Things Journal},
  year={2025},
  publisher={IEEE}
}

@inproceedings{andreychuk2025mapf,
  title={Mapf-gpt: Imitation learning for multi-agent pathfinding at scale},
  author={Andreychuk, Anton and Yakovlev, Konstantin and Panov, Aleksandr and Skrynnik, Alexey},
  booktitle={Proceedings of the AAAI Conference on Artificial Intelligence},
  volume={},
  pages={},
  year={2025}
}

@inproceedings{bhattacharyya2018multi,
  title={Multi-agent imitation learning for driving simulation},
  author={Bhattacharyya, Raunak P and Phillips, Derek J and Wulfe, Blake and Morton, Jeremy and Kuefler, Alex and Kochenderfer, Mykel J},
  booktitle={2018 IEEE/RSJ International Conference on Intelligent Robots and Systems (IROS)},
  pages={},
  year={2018},
  organization={IEEE}
}

@inproceedings{huang2023multi,
  title={Multi-agent decision-making at unsignalized intersections with reinforcement learning from demonstrations},
  author={Huang, Chang and Zhao, Junqiao and Zhou, Hongtu and Zhang, Hai and Zhang, Xiao and Ye, Chen},
  booktitle={IEEE Intelligent Vehicles Symposium (IV)},
  pages={},
  year={2023}
}

@article{song2018multi,
  title={Multi-agent generative adversarial imitation learning},
  author={Song, Jiaming and Ren, Hongyu and Sadigh, Dorsa and Ermon, Stefano},
  journal={Advances in neural information processing systems},
  volume={},
  year={2018}
}

@inproceedings{squires2018lfd,
  title={LfD Training of Heterogeneous Formation Behaviors.},
  author={Squires, William G and Luke, Sean},
  booktitle={AAAI Spring Symposia},
  year={2018}
}

@inproceedings{luke2010learn,
  title={Learn to behave! rapid training of behavior automata},
  author={Luke, Sean and Ziparo, Vittorio Amos},
  booktitle={Proceedings of adaptive and learning agents workshop at aamas},
  year={2010},
  organization={Citeseer}
}

@inproceedings{turco2024reducing,
  title={Reducing cognitive load in teleoperating swarms of robots through a data-driven shared control approach},
  author={Turco, Enrico and Castellani, Chiara and Bo, Valerio and Pacchierotti, Claudio and Prattichizzo, Domenico and Baldi, Tommaso Lisini},
  booktitle={2024 IEEE/RSJ International Conference on Intelligent Robots and Systems (IROS)},
  pages={},
  year={2024},
  organization={IEEE}
}

@inproceedings{losey2020controlling,
  title={Controlling assistive robots with learned latent actions},
  author={Losey, Dylan P and Srinivasan, Krishnan and Mandlekar, Ajay and Garg, Animesh and Sadigh, Dorsa},
  booktitle={2020 IEEE International Conference on Robotics and Automation (ICRA)},
  pages={},
  year={2020}
}

@article{fang2019survey,
  title={Survey of imitation learning for robotic manipulation},
  author={Fang, Bin and Jia, Shidong and Guo, Di and Xu, Muhua and Wen, Shuhuan and Sun, Fuchun},
  journal={International Journal of Intelligent Robotics and Applications},
  volume={},
  number={},
  pages={},
  year={2019},
  publisher={Springer}
}

@inproceedings{ross2011reduction,
  title={A reduction of imitation learning and structured prediction to no-regret online learning},
  author={Ross, St{\'e}phane and Gordon, Geoffrey and Bagnell, Drew},
  booktitle={International Conference on Artificial Intelligence and Statistics},
  pages={},
  year={2011}
}

@inproceedings{hoque2021thriftydagger,
  title={ThriftyDAgger: Budget-Aware Novelty and Risk Gating for Interactive Imitation Learning},
  author={Hoque, Ryan and Balakrishna, Ashwin and Novoseller, Ellen and Wilcox, Albert and Brown, Daniel S and Goldberg, Ken},
  booktitle={Conference on Robot Learning},
  year={2021}
}

@inproceedings{ross2010efficient,
  title={Efficient reductions for imitation learning},
  author={Ross, St{\'e}phane and Bagnell, Drew},
  booktitle={International conference on artificial intelligence and statistics},
  pages={},
  year={2010}
}

@inproceedings{farkhatdinov2008teleoperation,
  title={Teleoperation of multi-robot and multi-property systems},
  author={Farkhatdinov, Ildar and Ryu, Jee-Hwan},
  booktitle={2008 6th IEEE International Conference on Industrial Informatics},
  pages={},
  year={2008},
  organization={IEEE}
}

@article{patel2021transparency,
  title={Transparency in multi-human multi-robot interaction},
  author={Patel, Jayam and Ramaswamy, Tyagaraja and Li, Zhi and Pinciroli, Carlo},
  journal={arXiv preprint arXiv:2101.10495},
  year={2021}
}

@inproceedings{bettini2022vmas,
  title={Vmas: A vectorized multi-agent simulator for collective robot learning},
  author={Bettini, Matteo and Kortvelesy, Ryan and Blumenkamp, Jan and Prorok, Amanda},
  booktitle={International Symposium on Distributed Autonomous Robotic Systems},
  pages={},
  year={2022},
  organization={Springer}
}

@article{schulman2017proximal,
  title={Proximal policy optimization algorithms},
  author={Schulman, John and Wolski, Filip and Dhariwal, Prafulla and Radford, Alec and Klimov, Oleg},
  journal={arXiv preprint arXiv:1707.06347},
  year={2017}
}

@inproceedings{laskey2017dart,
  title={Dart: Noise injection for robust imitation learning},
  author={Laskey, Michael and Lee, Jonathan and Fox, Roy and Dragan, Anca and Goldberg, Ken},
  booktitle={Conference on robot learning},
  pages={},
  year={2017},
  organization={PMLR}
}

@inproceedings{kaduk2024one,
  title={From One to Many: How Active Robot Swarm Sizes Influence Human Cognitive Processes},
  author={Kaduk, Julian and Cavdan, M{\"u}ge and Drewing, Knut and Hamann, Heiko},
  booktitle={2024 33rd IEEE International Conference on Robot and Human Interactive Communication (ROMAN)},
  pages={},
  year={2024},
  organization={IEEE}
}

@inproceedings{mattson2025discovery,
  title={Discovery and Deployment of Emergent Robot Swarm Behaviors via Representation Learning and Real2Sim2Real Transfer},
  author={Mattson, Connor and Raveendra, Varun and Vega, Ricardo and Nowzari, Cameron and Drew, Daniel S and Brown, Daniel S},
  booktitle={Proceedings of the 24th International Conference on Autonomous Agents and Multiagent Systems},
  pages={},
  year={2025}
}

@article{rezeck2023hero,
  title={Hero 2.0: A low-cost robot for swarm robotics research},
  author={Rezeck, Paulo and Azp{\'u}rua, H{\'e}ctor and Correa, Mauricio FS and Chaimowicz, Luiz},
  journal={Autonomous Robots},
  volume={},
  number={},
  pages={},
  year={2023},
  publisher={Springer}
}

@article{mai2024inverse,
  title={Inverse factorized soft q-learning for cooperative multi-agent imitation learning},
  author={Mai, Tien and Nguyen, Thanh and others},
  journal={Advances in Neural Information Processing Systems},
  volume={},
  pages={},
  year={2024}
}

@article{fang2025multi,
  title={Multi-agent generative adversarial interactive self-imitation learning for AUV formation control and obstacle avoidance},
  author={Fang, Zheng and Chen, Tianhao and Shen, Tian and Jiang, Dong and Zhang, Zheng and Li, Guangliang},
  journal={IEEE Robotics and Automation Letters},
  year={2025},
  publisher={IEEE}
}

@article{aggravi2021decentralized,
  title={Decentralized control of a heterogeneous human--robot team for exploration and patrolling},
  author={Aggravi, Marco and Sirignano, Giuseppe and Giordano, Paolo Robuffo and Pacchierotti, Claudio},
  journal={IEEE Transactions on Automation Science and Engineering},
  volume={},
  number={},
  pages={},
  year={2021},
  publisher={IEEE}
}

@article{yu2022surprising,
  title={The surprising effectiveness of ppo in cooperative multi-agent games},
  author={Yu, Chao and Velu, Akash and Vinitsky, Eugene and Gao, Jiaxuan and Wang, Yu and Bayen, Alexandre and Wu, Yi},
  journal={Advances in neural information processing systems},
  volume={},
  pages={},
  year={2022}
}

@article{srinivasan2021fast,
  title={Fast multi-robot motion planning via imitation learning of mixed-integer programs},
  author={Srinivasan, Mohit and Chakrabarty, Ankush and Quirynen, Rien and Yoshikawa, Nobuyuki and Mariyama, Toshisada and Di Cairano, Stefano},
  journal={IFAC-PapersOnLine},
  volume={},
  number={},
  pages={},
  year={2021},
  publisher={Elsevier}
}

@article{chen2023transformer,
  title={Transformer-based imitative reinforcement learning for multirobot path planning},
  author={Chen, Lin and Wang, Yaonan and Miao, Zhiqiang and Mo, Yang and Feng, Mingtao and Zhou, Zhen and Wang, Hesheng},
  journal={IEEE Transactions on Industrial Informatics},
  volume={},
  number={},
  pages={},
  year={2023},
  publisher={IEEE}
}

@article{li2019two,
  title={A two-stage imitation learning framework for the multi-target search problem in swarm robotics},
  author={Li, Jie and Tan, Ying},
  journal={Neurocomputing},
  volume={},
  pages={},
  year={2019},
  publisher={Elsevier}
}

@article{queralta2020collaborative,
  title={Collaborative multi-robot search and rescue: Planning, coordination, perception, and active vision},
  author={Queralta, Jorge Pena and Taipalmaa, Jussi and Pullinen, Bilge Can and Sarker, Victor Kathan and Gia, Tuan Nguyen and Tenhunen, Hannu and Gabbouj, Moncef and Raitoharju, Jenni and Westerlund, Tomi},
  journal={Ieee Access},
  volume={},
  pages={},
  year={2020},
  publisher={IEEE}
}

@article{scherer2020multi,
  title={Multi-robot persistent surveillance with connectivity constraints},
  author={Scherer, J{\"u}rgen and Rinner, Bernhard},
  journal={IEEE Access},
  volume={},
  pages={},
  year={2020},
  publisher={IEEE}
}

@inproceedings{gregory2016application,
  title={Application of multi-robot systems to disaster-relief scenarios with limited communication},
  author={Gregory, Jason and Fink, Jonathan and Stump, Ethan and Twigg, Jeffrey and Rogers, John and Baran, David and Fung, Nicholas and Young, Stuart},
  booktitle={Field and service robotics: results of the 10th international conference},
  pages={},
  year={2016},
  organization={Springer}
}

@article{bolu2021adaptive,
  title={Adaptive task planning for multi-robot smart warehouse},
  author={Bolu, Ali and Kor{\c{c}}ak, {\"O}mer},
  journal={Ieee Access},
  volume={},
  pages={},
  year={2021},
  publisher={IEEE}
}

@inproceedings{hoque2023fleet,
  title={Fleet-dagger: Interactive robot fleet learning with scalable human supervision},
  author={Hoque, Ryan and Chen, Lawrence Yunliang and Sharma, Satvik and Dharmarajan, Karthik and Thananjeyan, Brijen and Abbeel, Pieter and Goldberg, Ken},
  booktitle={Conference on Robot Learning},
  pages={368--380},
  year={2023},
  organization={PMLR}
}

% \clearpage
% \subfile{appendix}

\end{document}